\documentclass[letterpaper]{article} 
\usepackage[draft]{aaai25}  
\usepackage{times}  
\usepackage{helvet}  
\usepackage{courier}  
\usepackage[hyphens]{url}  
\usepackage{graphicx} 
\usepackage{natbib}  
\usepackage{caption} 
\usepackage{amsmath,amssymb}
\usepackage[utf8]{inputenc} 
\usepackage[T1]{fontenc}    
\usepackage{url}            
\usepackage{booktabs}       
\usepackage{amsfonts}       
\usepackage{nicefrac}       
\usepackage{microtype}      
\usepackage{xcolor}         
\usepackage{subcaption,moreverb,marvosym}
\usepackage{algpseudocode}
\def\f{\frac}\def\l{\left}\def\r{\right}

\title{The Ungrounded Alignment Problem}
\author{
  Marc Pickett\textsuperscript{\includegraphics[width=0.01\textwidth]{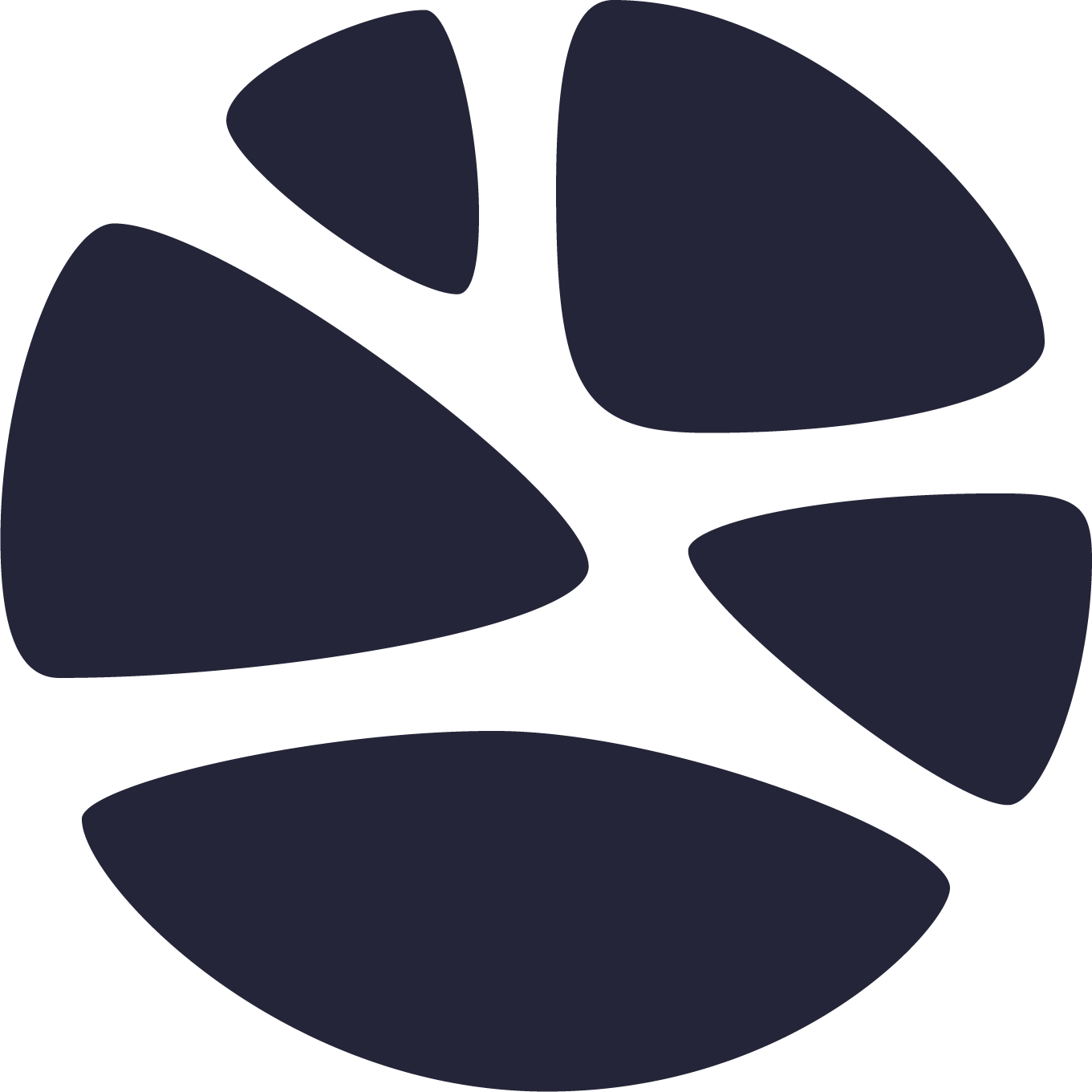}},
  Aakash Kumar Nain\textsuperscript{\includegraphics[width=0.01\textwidth]{./figures/emergence_symbol.png}},
  Joseph Modayil\textsuperscript{$\mho$},
  Llion Jones\textsuperscript{\includegraphics[width=0.01\textwidth]{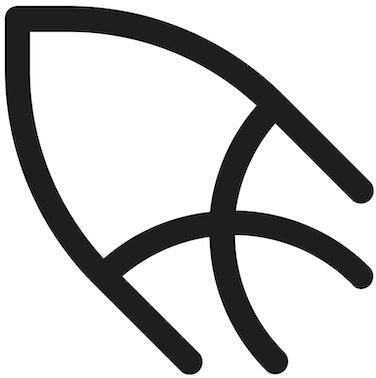}}
}
\affiliations{
  \textsuperscript{\includegraphics[width=0.01\textwidth]{./figures/emergence_symbol.png}}{\rm Emergence AI}
  \\ \textsuperscript{$\mho$}{\rm Openmind Research Institute}
  \\ \textsuperscript{\includegraphics[width=0.01\textwidth]{./figures/sakana_symbol.png}}{\rm Sakana AI, Tokyo}
  \\ \texttt{mpickett@emergence.ai}
}

\begin{document}
\maketitle 
\begin{abstract} 
Modern machine learning systems have demonstrated substantial abilities with methods that either embrace or ignore human-provided knowledge, but
combining benefits of both styles remains a challenge.
One particular challenge involves designing learning systems that exhibit built-in responses to specific abstract stimulus patterns, yet are
still plastic enough to be agnostic about the modality and exact form of their inputs.
In this paper, we investigate what we call \emph{The Ungrounded Alignment Problem}, which asks \emph{How can we build in predefined knowledge in
a system where we don't know how a given stimulus will be grounded?}
This paper examines a simplified version of the general problem, where an unsupervised learner is presented with a sequence of images for the
characters in a text corpus, and this learner is later evaluated on its ability to recognize specific (possibly rare) sequential patterns.
Importantly, the learner is given no labels during learning or evaluation, but must map images from an unknown font or permutation to its
correct class label.
That is, at no point is our learner given labeled images, where an image vector is explicitly associated with a class label.
Despite ample work in unsupervised and self-supervised loss functions, all current methods require a \emph{labeled} fine-tuning phase to map the
learned representations to correct classes.  Finding this mapping in the absence of labels may seem a fool's errand, but our main result
resolves this seeming paradox.  We show that leveraging only letter bigram frequencies is sufficient for an unsupervised learner both to
reliably associate images to class labels and to reliably identify trigger words in the sequence of inputs.
More generally, this method suggests an approach for encoding specific desired innate behaviour in modality-agnostic models.
\end{abstract} 

\section{Introduction} 
Both biological and artificial systems benefit from general learning, since adaptability is a key to robust success.
Artificial systems, such as Transformers, have shown that the same circuitry can be used for language \cite{vaswani2023attention}, vision
\cite{dosovitskiy2021image}, and other modalities \cite{he2024sonarnet}, with a main differentiator being the data fed to these systems.
For mammalian brains, the Mountcastle hypothesis proposes that the neocortex is a general learning system \cite{mountcastle1997columnar}. The
same cortical circuitry that performs high-level planning also performs seeing and hearing.  Sur and Rubenstein \cite{sur2005patterning} even
suggest that a newborn ferret's auditory cortex can learn to ``see'' given visual instead of aural input.

Nevertheless, innate instincts serve an essential role for a species' survival.  An anecdotal example is a beaver raised in human captivity
since infancy that built a ``dam'' inside its human owners' home using household
items\footnote{\url{https://youtu.be/-ImdlZtOU80?si=PlyfhUK6NCYU7isK}}.  This beaver had never been instructed on how to build a
dam, yet it had a drive to do so.
One possibility for how this drive is genetically encoded is that a special module in the beaver's brain is dedicated to dam building.  This
module ``hard codes'' the neural structure all the way from grounding in visual and auditory signals to motor control, essentially saying ``When
you observe this pattern, take these actions''.  The problem with the approach is this module would presumably break if we were to reroute a
baby beaver's optic nerve to its auditory cortex.  Or if we were to permute the baby beaver's retinotopic mapping, essentially changing the
``pixels'' that travel along the optic nerve \cite{sperry1943effect}.  Further, the ''actions'' used in such encodings would also need grounding
to environmentally-appropriate affordances, as the baby beaver dwelt in a human house that did not contain the branches and mud that is more
commonly used in the construction of beaver dams.

The problem we are investigating in this paper is how can we achieve similar innate instincts in artificial neural networks?  More specifically,
we ask: \emph{How can we build in specific desired concepts in a system where we don't know how a given stimulus will be grounded?}  Each time a
neural network is trained from scratch, the internal representations that it learns will be different (due to random initialization) and thus we
are not able to simply build in a module that detects a given stimulus in any given representation.  Though there is some evidence that models
do learn similar representations (up to permutations) when trained on the same data even after random initialization \cite{entezari2022role}, it
seems that existing literature has very little to say on how we might solve the above problem.  We call this \emph{The Ungrounded Alignment
Problem}.

More pragmatically, it could be useful to have a module that, when attached to a robot with uninterpreted sensors \cite{pierce1997map}, would
give the robot an innate drive to pick up trash, for example.  Such a module would need to allow the robot to define and detect ``trash''
independent of its specific modality.  In this paper, we investigate a first step of this process, merely detecting specific high-level concepts
without explicit grounding at design-time.

We view our main contribution as introducing an interesting problem (Ungrounded Alignment) that seems to have a lack of solutions in the
literature.  Our secondary contributions are a formalization of a simplified version of this problem and a demonstration of its
%
solution\footnote{Our code is available at \url{https://github.com/EmergenceAI/babybeaver}.}.
More specifically, our contributions are:
\begin{itemize} 
\item In Section \ref{section:babybeaver}, we formalize the Ungrounded Alignment Problem, and provide the specific instance of {\tt fnord}
  detection: The problem of learning to detect a ``trigger'' sequence of images representing specific characters (e.g., {\tt f}-{\tt n}-{\tt
    o}-{\tt r}-{\tt d}) without using labels for either characters or the trigger sequence, where the characters are in a font that is unknown
  at design time.  We argue this instantiation captures the core of the more abstract problem (namely, grounding specific high level concepts in
  uninterpreted sensors without relying on labels during training).
\item We argue that usual unsupervised methods are insufficient to solve this, demonstrating the results of clustering in Appendix
  \ref{appendix:clustering}, and that even taking into account single character frequencies (unigrams) alone is insufficient for this task
  (Section \ref{subsec:alignmentloss}).
\item In Section \ref{section:solution} we propose a solution for the formalized problem, which, at its core, uses a simple bigram ``alignment''
  loss function described in Subsection \ref{subsec:alignmentloss}.
\item In Section \ref{section:results}, we show that our solution achieves over 99\% test accuracy on our {\tt fnord} detection task (vs.\ 50\%
  random or max-class), effectively solving our introduced challenge.
\item In Subsection \ref{subsec:characters} we show that our model achieves 82\% test accuracy on single-character classification for permuted
  Extended-MNIST and 23\% on a 26-class subset of permuted CIFAR100 (vs.\ 3.8\% random accuracy for 26 character classes) \emph{without labels or
  finetuning}.
\end{itemize} 

\section{Specifying the Problem} 
\label{section:babybeaver}

To create a formal simplified version of the Ungrounded Alignment Problem, we make some assumptions to simplify the problem setup while
preserving our core concerns.
\begin{enumerate} 
\item The learner's experience is completely unsupervised.  At no point does the learning or evaluation process supply explicit labels (from
  images to class labels).
\item There are environmental invariants that we assume to be stable from design time to deployment time.  As an example, we could assume that,
  though a trash-picking robot's sensors may change dramatically, the overall dynamics of its environment are relatively steady.
\item Solving a classification problem \emph{without relying on labels or a specific sensory configuration} is sufficient to address the core
  problem, which is how specific concepts can be innately encoded without any access to grounding, labels, or feedback during deployment.  If a
  robot learns concepts like ``trash'' and``pick up'', we hope it's not too far a stretch to imagine that it can be internally rewarded for
  ``picking up trash''.
\end{enumerate} 

\begin{figure}[ht]
  \centering
  \begin{minipage}{0.49\textwidth}
    \centering
    \includegraphics[width=0.95\textwidth]{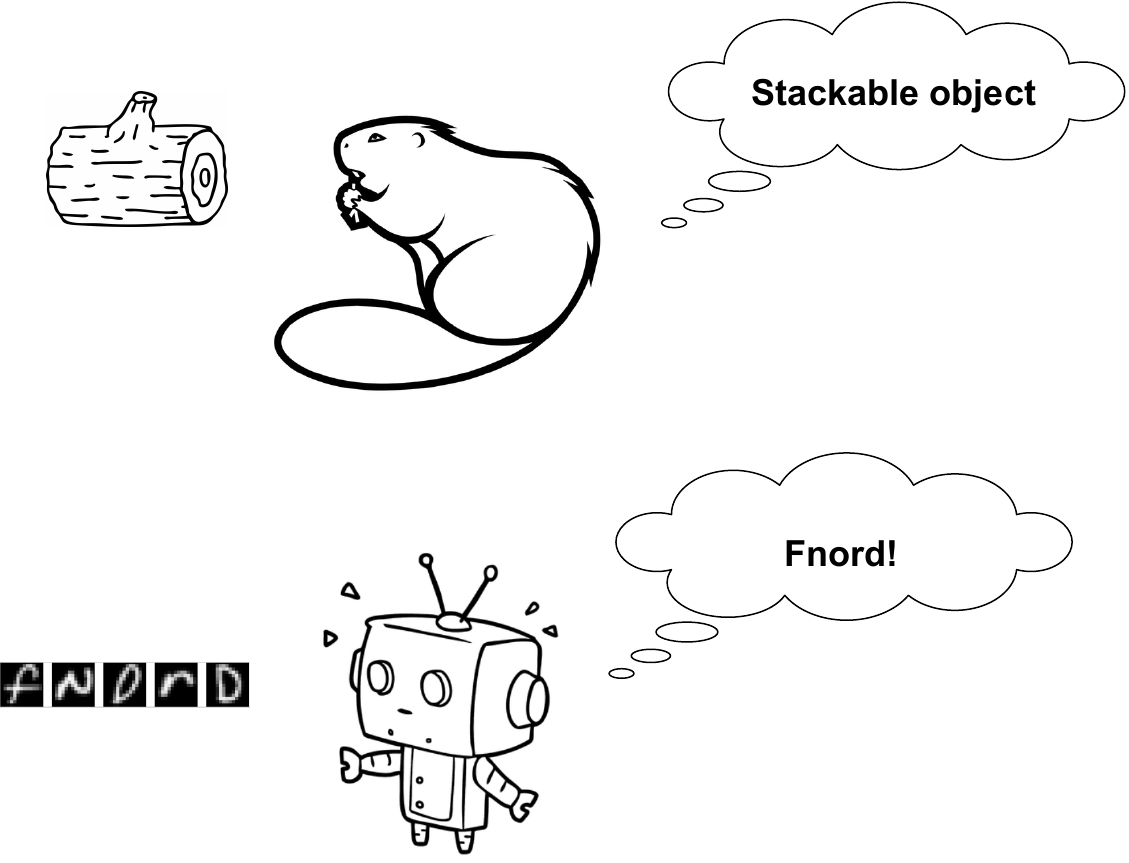}
    \caption{An analogy of our simplified version: A beaver must be able to connect abstract concepts, like ``stackable objects'' to sensors
      without labels or modality-specific wiring.  In our formalized problem, the model's task is to recognize instances of particular character
      sequences while being modality agnostic (before training) and without the use of labels.}
    \label{figure:analogy}
  \end{minipage}
\end{figure}

Given these assumptions, we propose a simplified but well-defined instance of The Ungrounded Alignment Problem with {\tt fnord} detection (with
an analogy shown in Figure \ref{figure:analogy}).  The core issue in both the full and simplified version is grounding specific concepts, like
``movable object'', using inputs whose precise encoding is unknown at design-time.

\subsection{The Ungrounded Alignment Problem}

We start by presenting a formal version of the problem here, and then provide a specific instantiation with fnords that we use in experiments
below.

We develop a model for eventual deployment with a training and test phase.  For the training phase, the model receives a sequence of high
dimensional observations $X= x_1, x_2, x_3, \ldots,$.  Each observation $x_i$ is drawn from a high-dimensional subspace, $x_i \in \mathbb{X}
\subseteq \mathbb{R}^D$. Each observation $x$ corresponds to a letter $\sigma$ from a finite alphabet $\Sigma=\{
\sigma_1,...,\sigma_{|\Sigma|}\}$, where the correspondence is provided by a function $\phi: \mathbb{X}\rightarrow \Sigma$.  The specific
function $\phi$ is unknown to the algorithm designer and the model, but the algorithm designer has access to the finite alphabet $\Sigma$ and
they may know other side-channel information about the environment.

In the test phase, there is a special trigger word, $T=t_1 \ldots t_n$, where $t_i\in \Sigma$.  The model is given a set of length-$n$ words, of
which a known percentage are instances of the trigger word.  Whenever a subsequence of the observation stream matches the trigger, namely
$\phi(x_{i+1})=t_1 \wedge \ldots \wedge \phi(x_{i+n})=t_n$, then the model should immediately emit an innate response $\rho_1$, otherwise it
should emit a null response, $\rho_0$.  Importantly, no feedback or labels are given to the model as to whether the response is correct or
incorrect, so the model cannot adapt to feedback.  However, high accuracy is important for the model to be suitable for its environment, and
thus it is the measure of success for the algorithm.

\subsection{An instantiation with ``fnord''} 

\begin{figure*}[ht]
  \centering
  \includegraphics[width=0.95\textwidth]{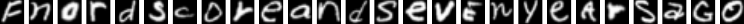}
  \\ \hphantom{.}
  \\
  \includegraphics[width=0.95\textwidth]{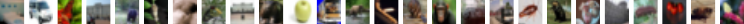}
  \caption{{\bf Above:} An example sequence of images representing the string {\tt fnordscoreandsevenyearsago} in the EMNIST ``font''.  Note that
    upper and lower case character forms are used interchangeably.
    \\
    {\bf Below:} The same string in the ``CIFAR26 font''.  In this ``font'', {\tt a} is represented by images of apples, {\tt f} by beds, {\tt
      n} by buses, etc., assigning the first 26 CIFAR100 classes to letters.  In our experiment, we simply ordered the classes alphabetically,
    so {\tt b} is ``aquarium fish''. (See Appendix \ref{appendix:cifar} for more details.)
  }
  \label{figure:fnordscore}
\end{figure*}

In this setup, we assume a passive unsupervised learning system.  There are no labels or actions.  The model is given a stream of fixed-length
vectors, each corresponding to one of 26 English letters.  The sequences are from common English texts.  We assume that the \emph{character
sequence} distribution is stable for all worlds, but the \emph{sensor modality} distribution is unknown at design time.  The system is given no
other information from the environment.  We consider our system a success if after ``training'' there is a node that is active if and only if
the system has just seen a sequential (innate) pattern whose exact grounded form must be learned.  In our examples we use the ``trigger'' word
{\tt fnord}, meaning that the model will need to detect a sequence of images of the letters {\tt f}, {\tt n}, {\tt o}, {\tt r}, then {\tt d} in
a font that is unknown, in an unknown representation mapping, at design time.  In our experiments, we use images randomly selected from permuted
EMNIST \cite{cohen2017emnist} and from permuted CIFAR100 \cite{krizhevsky2009learning}, as shown in Figure \ref{figure:fnordscore} (the shown
images have been un-permuted and unflattened for interpretability).  We use the sequence {\tt fnord} in our examples, but the trigger may be any
specific string\footnote{The term {\tt fnord} appears in the \emph{Illuminatus!}  trilogy \cite{illuminatus}, and is a term that people in the
story innately fear.}.  The term {\tt fnord} is uncommon, so our method should not depend on seeing the target sequence during training.
An accurate ``{\tt fnord} detector'' could then be used by an external process for an appropriate response.

The letters that the vectors represent have a 1:1 correspondence with the 26 letters of the English alphabet, but the model isn't given labels
or any explicit knowledge whether two vectors represent the same letter\footnote{If we assume we're told whether two vectors represent the same
letter, then the problem is a fairly trivial substitution cipher decoding.}.  Each letter vector is a fixed dimension $D$ (784 for EMNIST, 3,072
for CIFAR), but we don't know what this dimension is beforehand.

More formally, the model is given a stream of (possibly permuted) handwritten images of single letters $X = x_1, x_2, x_3, \cdots$, where each
$x_i$ is of dimension $D$.  (A short example sequence is shown in Figure \ref{figure:fnordscore}.)  Each letter represents one of $\l|\Sigma\r|
= 26$ classes from an alphabet $\Sigma$, but the class labels are not given.

The characters represent a sequence of text (English Wikipedia in our case, from {\tt wikipedia20220301.en}, uncased, with skipped
non-alphabetic characters).  After training the encoder on $X$ (only the inputs, no labels), the model's task is to detect instances of a
\emph{trigger} word $T = t_1, t_2, t_3, \cdots t_n$ (e.g., {\tt fnord}) in a new stream of test input $X'$, which is drawn from the same
distribution as $X$ (both the letter images and the sequences are drawn from held-out data).  Note that the original stream $X$ may or may not
contain instances of the trigger word.  The model's final score is classification accuracy given a balanced test set, including 10,000 examples
each of the trigger word and non-trigger sequences of length $\l|T\r|$ sampled from the held-out data.

For our experiments, for each character, we randomly sampled corresponding images from EMNIST \cite{cohen2017emnist}, which contains both upper
and lower case versions of each character interchangeably.  EMNIST does not have characters for spaces or punctuation, so we removed these from
the text.
Since we want our system to be modality-agnostic (with the constraint that the modality is a sequence of fix-width vectors of a known
dimension), the images are flattened and permuted to prevent reliance on modality-specific knowledge (similar to that introduced in
\cite{goodfellow2015empirical}), so our model can't make use of convnets because they rely on spatial assumptions.  We also repeated our
experiments using images from CIFAR100, using the first 26 classes (sorted alphabetically by label name).  (We call this ``font'' CIFAR26.  See
Appendix \ref{appendix:cifar} for more detail.)  The only other change from EMNIST to CIFAR was increasing the input dimension from 784 to
3,072.
For EMNIST our train and test sets have roughly 3,800 and 950 examples of each character, respectively.  For CIFAR26, these figures are 380 and
110.  During training, the model sees a sequence of 2M input vectors, representing 2M characters.  Note that, because of the relatively low
number of total character images, the character images will be seen multiple times in the training sequence, which could lead to over-fitting.
However, the test character images are not seen during training, which will penalize solutions that overfit.

\section{A Solution} 
\label{section:solution}

The crux of our model is training an \emph{unsupervised} character \emph{encoder} to map a letter image $x_i$ to its correct class label (or to a
probability distribution over all the classes).
The encoder is to be trained from scratch \emph{without using labels during training}.
With such an encoder, detecting instances of the trigger word $T$ becomes trivial.  But how can we map to the correct class label when we are
never given labels during training?  We can't rely on our knowledge of images because the images may be permuted or in an absurd new font.  Our
problem is non-trivial because the system needs to simultaneously learn which letter map to which and which letters are the same as others.

If we could \emph{cluster} the images into 26 clusters with a reliable one to one correspondence between cluster and label, then our mapping
problem would be reduced to easily solving a simple cryptographic substitution cipher \cite{ramesh1993automated}.  However, depending on the
font, clustering alone rarely produces such a correspondence, with in-class similarity often being less than cross-class similarity.  For
example, the pixel distance between images of lowercase {\tt a} and {\tt o} is often smaller than that for two images of {\tt z} which has
crossed and uncrossed varieties.  (In Appendix \ref{appendix:clustering}, we show the results of K-means clustering on raw EMNIST images and its
lack of 1:1 correspondence between clusters and characters.)  For similar reasons, virtually any loss function that learns solely on images of
individual letters is unlikely to yield a clustering that is one-to-one with the letter's true labels.  (In Section \ref{section:results}, we
show the poor performance of unigram models, even when taking into account known character frequencies.)

\begin{figure}[ht]
      \begin{minipage}{0.49\textwidth}
    \centering
    \includegraphics[width=0.95\textwidth]{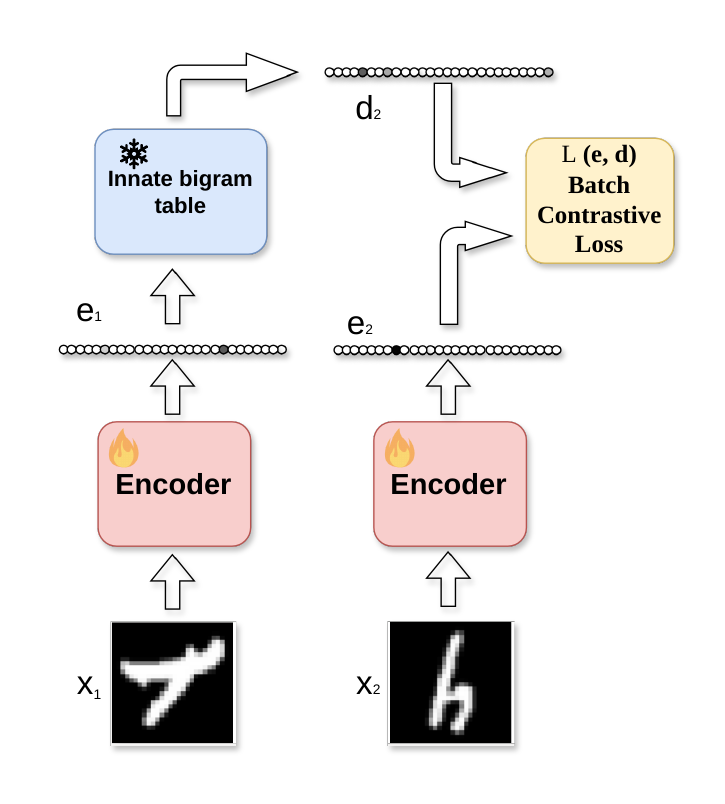}
    \caption{The loss process.  The encoder is shared for both input images, and is trained from scratch using the bigram probability table and
      batch contrastive loss.  Note that no labels are used in this process.  The bigram table is ``innate'' and fixed.}
    \label{figure:batchloss}
  \end{minipage}
\end{figure}

This may seem paradoxical: \emph{unsupervised} training an encoder from scratch to give correct class \emph{labels} (without seeing the labels
during training).  However, the key to our approach (inspired by the ``conceptual web'' account of concepts \cite{goldstone2002using}) is to
exploit ``innate'' knowledge of the \emph{relationships} among the abstract concepts\footnote{From the viewpoint of raw pixels, we consider
an image's letter classification to be an ``abstract'' concept.}.  Specifically, we provide our model with hard-coded knowledge of the bigram
distribution for characters in English text.  Our final model uses a known bigram distribution $Bi\l(y|x\r)$ that simply returns the probability
that character $y$ immediately follows character $x$.  (E.g., $Bi\l(\mathtt{h}|\mathtt{t}\r) \approx .14$, or the probability of {\tt h}
following {\tt t} is 14\%.)  The bigram distribution is both ``innate'' and fixed, meaning the distribution never changes while training the
encoder.  In our experiments, this is a simple lookup table formed by counting bigrams from the Wikipedia text.
Our encoder is trained from scratch using the frozen bigram probability table using a batch contrastive loss as shown in Figure
\ref{figure:batchloss} and described below.

\subsection{The ``alignment'' loss function} 
\label{subsec:alignmentloss}

Given 1.\ an encoder $E$ that maps images to class probabilities, 2.\ our bigram distribution $Bi$, and 3.\ two sequential letter images $x_i$ and
$x_{i+1}$, we define a loss function that compares the agreement between a.\ the classes for the last image as directly predicted by the
encoder, and b.\ the classes predicted by the bigram table (given the class distributions from the encoder for the first image).

More formally, let $e_i = E\l(x_i\r)$ be the class probabilities predicted by the encoder for $x_i$, such that $e_{i,j}$ is the encoder's
predicted probability that $x_i$ represents character $j$.  The bigram's predicted probability that $x_{i+1}$ is character $y$ is $d_{i+1, y}$:
\begin{displaymath} 
d_{i+1, y} \equiv P_{Bi}\l(y|e_{i}\r) = \sum_{x \in \Sigma} Bi\l(y| e_{i, x}\r)
\end{displaymath} 

Ideally, the two distributions $d_i$ (output of the bigram table) and $e_i$ (output of the encoder) will match.
We use a batch-contrastive loss that measures how well we can match specific items of our batch given both predicted distributions $d$ and $e$.
To do this, we seek to maximize the amount of information (entropy) that $d$ gives us about $e$.  We already know that $d_i$ and $e_i$ should be
encodings to predict the same letter.  More specifically, as derived in Appendix \ref{appendix:contrastloss}, this loss is:
\begin{equation} 
  \label{eqn:theloss}
  \mathcal{L}\l(e, d\r) = -\f{1}{\l|B\r|} \sum_{i=1}^{\l|B\r|} \log\l(\sum_{j=1}^{\l|\Sigma\r|} \f{d_{i,j} e_{i,j}}{\sum_{k=1}^{\l|B\r|} e_{k,j}}\r)
\end{equation} 

Note that single character (unigram) approaches alone, such as simply matching a single character distribution, $\mathcal{L}_c \l(e\r) =
KL\l(c||e\r)$ where $c_i$ is the prior probability for character $i$, are insufficient for disambiguating characters with similar frequencies.
For example, there is nothing to break the symmetry between {\tt l} and {\tt h}, which have nearly identical priors (prob $\approx$ 0.036725)
when computed by counting the character frequencies in the Wikipedia text.  We show empirical results for such models in Section
\ref{section:results}.
Another possible approach is simply to match the batch distribution for bigrams using KL from the ``true'' distribution similar to the approach
used by \cite{sutskever2015principled}.  Specifically, this loss is $\mathcal{L}_{kl}\l(e, d\r) = \sum d \log\l(d/e\r)$.  In section
\ref{section:results}, we show that this led to worse performance especially for EMNIST.

\subsection{Model and Optimization} 

\begin{figure}[ht]
  \centering
  \includegraphics[width=0.48\textwidth]{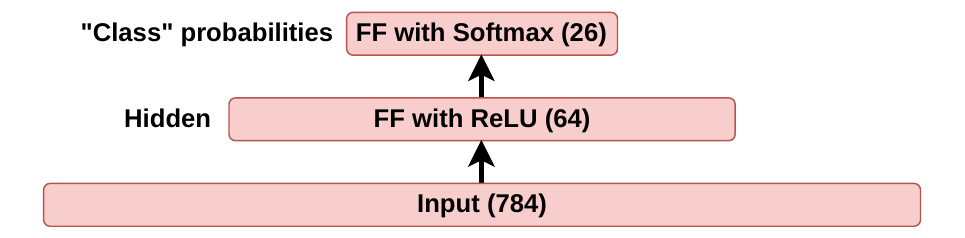}
  \caption{The encoder.}
  \label{figure:theencoder}
\end{figure}

Our encoder maps an input image to one of 26 classes.  It is a simple two layer feed fnorward network with biases and ReLU activations and
softmax output: 784 inputs to 64 hidden units to 26 outputs (for CIFAR, it was 3,072 inputs).  The encoder (Figure \ref{figure:theencoder}) has
just over 50K parameters (this is EMNIST, after all), and 200K parameters for CIFAR.  The encoder is shared for encoding each image during
optimization in our bigram model (Figure \ref{figure:batchloss}).

Though our encoder is simple, optimization was less so because of local optima (see Appendix \ref{appendix:optimal}).  To mitigate issues of
local optima, we used 64 random restarts and selected the model with the lowest training loss.

Given the encoder, we hard-coded a simple {\tt fnord} detector that simply sums the log probabilities that $x_i$ is the first letter of our
trigger $T_1$ (i.e., {\tt f}), $x_{i+1}$ is {\tt n}, $\cdots$, and $x_{i+4}$ is {\tt d}.  The detector reports ``True'' iff this sum over all
letters in the sequence is above a fixed threshold $\theta$.  The threshold is set by relying on an \emph{innate prior} of the trigger's
probability $P\l(T\r)$.  Specifically, at test time, we set the threshold $\theta$ such that the ratio of ``True'' is near $P\l(T\r)$ for the
test set.  One observation is that this threshold is significantly different for EMNIST and CIFAR, that is the best threshold $\theta$ is
$\approx 10^{-3}$ per character for EMNIST vs.\ $10^{-8}$ for CIFAR.  Note that setting the threshold does not depend on a teacher saying
\emph{which} sequences are the target, but just how often the target will appear.  It's important that, apart from the encoder, this detector
itself is innate (not tuned during learning), since the model will have no labels of whether or not a {\tt fnord} is present.

At a high level, the algorithm is summarized as follows.
\begin{algorithmic}[1]
\hrule
\item Initialize K=64 independent models (structure in Fig.\ \ref{figure:theencoder}).
\item Collect $N_0=2^{}$ samples, train all models with loss $\mathcal{L}_f$.
\item Evaluate model $m$ with lowest training loss.
\item Set $P\l(x_1 \cdots x_n\r) \equiv \prod_{i=1}^n P_m(x_i = t_i)$, with probabilities $P_m$ from the model.
\item Set response to $\rho_1$ when $P\l(x_1 \cdots x_n\r) > \theta$ (where $\theta$ is set using the prior $P\l(T\r)$, as described above).
\hrule
\end{algorithmic}
In our experiments, we formed our training set of $2^{20}$ pairs ($\approx$ 2M characters) into a single batch.  Our bigram-based models saw the
2M characters as a batch of 1M pairs, while the unigram-based models viewed the batch as 2M individual characters.  For robustness, we report
the mean and standard error of results for 10 different training / test splits.

\section{Results} 
\label{section:results}

Our primary result is that our model successfully detected {\tt fnord}.  Given 10,000 each of 5-character sequences of the trigger {\tt fnord}
and other non-trigger strings sampled from the Wikipedia test set (both with images from a held-out test set), our top model attained over 99\%
accuracy on EMNIST and 85\% on CIFAR as shown in Table \ref{table:fnord}, where random accuracy on this balanced test set is 50\%.

We also tested the robustness of the approach for other trigger words besides {\tt fnord}.  For each length from 2 characters to 11 characters
(inclusive), we sampled both 100 random strings from $\Sigma$ and 100 strings from the Wikipedia corpus as triggers, resulting in 2000 trigger
words.  For each trigger, we repeated the {\tt fnord} detection process above replacing {\tt fnord} with the new trigger and limiting the
samples to 100 instead of 10,000.  Some of the triggers, such as {\tt mdkjbebmrf}, did not appear at all in the training text at all, while
others were common.  We show the average test accuracy for detecting these triggers for EMNIST and CIFAR in Table \ref{table:fnord}.  (In
Appendix \ref{appendix:triggerlen}, we also show that longer triggers are easier to detect than shorter triggers.)

Finally, we also show the failure of two ``unigram'' models, where the loss used is based on simple KL-divergence ($\mathcal{L}_{kl} \l(e\r) =
KL\l(c||e\r)$, using $c$ and $d$ as defined in Section \ref{subsec:alignmentloss}), or also including a batch contrastive loss term
($\mathcal{L}_c \l(e\r) = KL\l(c||e\r) + \mathcal{L}\l(e, e\r)$, where $\mathcal{L}\l(e, e\r)$ is defined in Equation \ref{eqn:theloss}).  The
former loss can be minimized by simply ignoring the input and always generating $c$ for each input.  The contrastive term prevents this mode
collapse, but is still insufficient (perhaps because of symmetries of single-character frequencies).

\begin{table*}[ht]
  \centering
  \begin{tabular}{|l|c|c|c|c|}
    \hline
    \multicolumn{1}{|c}{Method} & \multicolumn{2}{|c|}{EMNIST-based font} & \multicolumn{2}{|c|}{CIFAR-based font} \\
    \cline{2-5}
    & Fnord Acc & 2K Trigger Acc & Fnord Acc & 2K Trigger Acc \\
    \hline
    Bigram Contrastive   & {\bf 99.88\% $\pm$ 0.01\%} & {\bf 99.09\% $\pm$ 0.01\%} & {\bf 85.49\% $\pm$ 1.59\%} & {\bf 83.73\% $\pm$ 1.22\%} \\
    Bigram KL-Divergence & 98.18\% $\pm$ 0.31\% & 97.71\% $\pm$ 0.12\% & 81.90\% $\pm$ 0.65\% & 77.41\% $\pm$ 1.11\% \\
    Unigram Contrastive  & 64.12\% $\pm$ 2.33\% & 56.41\% $\pm$ 2.24\% & 56.08\% $\pm$ 3.37\% & 53.51\% $\pm$ 1.08\% \\
    Unigram KL-Divergence& 69.08\% $\pm$ 0.62\% & 61.40\% $\pm$ 0.60\% & 47.47\% $\pm$ 0.62\% & 49.16\% $\pm$ 0.22\% \\
    \hline
  \end{tabular}
  \caption{Test accuracy for the best {\tt fnord} detector (of 64 seeds) for each loss type for both the EMNIST and CIFAR ``fonts''.  Random
    accuracy is 50\%.  We also show the average accuracy across 2,000 different trigger words in the ``2K Trigger Acc'' columns.}
  \label{table:fnord}
\end{table*}

\subsection{Character Classification} 
\label{subsec:characters}

We also evaluated the \emph{untuned} classification accuracy for our encoders on the EMNIST and CIFAR26\footnote{We define CIFAR26 to be the
subset of CIFAR100, but limited to only the first 26 classes, sorted alphabetically.} datasets, using the highest logit as the predicted class.
Note that the classes in the test set are evenly balanced --there are as many {\tt q}s as {\tt e}s--, so random is 1/26 or 3.85\%.  Our
encoders' losses bias them so that they will more accurately classify higher-frequency characters.  We achieved 82.14\% accuracy on the EMNIST
test set and 23.08\% on CIFAR26 (vs.\ 3.85\% for random), as shown in table \ref{table:emnist}.  While this is significantly worse than
state-of-the-art EMNIST and CIFAR100 (both $>$ 96\%) models\footnote{\cite{jeevan2024wavemixresourceefficientneuralnetwork} reports 95.96\%
accuracy on EMNIST, while \cite{foret2021sharpnessaware} reports over 96\% test accuracy for CIFAR100.  Having fewer classes, the results for
CIFAR26 would likely be even higher using the latter approach.}, it is still a remarkable result given that our model was given neither labels
nor structural information about the characters (i.e., the pixels were permuted).
We also trained an ``Oracle'' model using the same simple architecture used by our model (shown in Figure \ref{figure:theencoder}).  Remarkably,
our model achieved comparable performance for balanced EMNIST classification as this ``cheating'' model, though trained with a different
objective.  The poor performance of the unigram-based models is consistent with the results in Table \ref{table:fnord}.

\begin{table*}[ht]
  \centering
  \begin{tabular}{|l|l|l|}
    \hline
    \multicolumn{1}{|c}{Method} & \multicolumn{1}{|c|}{EMNIST} & \multicolumn{1}{|c|}{CIFAR26} \\
    \hline
    Bigram Contrastive (ours)   & {\bf 82.14\% $\pm$ 0.42\%} & {\bf 23.80\% $\pm$ 1.21\%} \\
    Bigram KL-Divergence & 66.71\% $\pm$ 0.43\% & 21.08\% $\pm$ 0.45\% \\
    Unigram Contrastive  & \hphantom{0}4.47\% $\pm$ 0.59\%  & \hphantom{0}4.74\% $\pm$ 0.35\% \\
    Unigram KL-Divergence& \hphantom{0}4.39\% $\pm$ 0.25\%  & \hphantom{0}3.82\% $\pm$ 0.22\% \\
    Random  / Max class                  & \hphantom{0}3.85\%   & \hphantom{0}3.85\% \\
    ``Oracle'' trained \emph{with labels}& 82.26\%   & 33.36\% \\
    \hline
  \end{tabular}
  \caption{Test classification accuracy of the best encoder (of 64 seeds each) on EMNIST and on the first 26 classes of CIFAR.  The ``Oracle''
    model is the same encoder architecture, but trained with labels using cross-entropy.  For fair comparison, we report the best of 64 seeds
    for the Oracle.}
  \label{table:emnist}
\end{table*}

\section{Related Work} 

Our work builds on a rich literature of unsupervised loss functions, self-supervised learning, and representational alignment.

In general, both unsupervised and self-supervised methods learn from unlabeled data, with the goal of learning representations that will
presumably be useful for downstream tasks like classification, generation, or control \cite{schmarje2021survey}.  For example, JEPA
\cite{assran2023selfsupervised}, Bootstrap Your Own Latent \cite{grill2020bootstrap}, and others \cite{tomasev2022pushing} learn representations
that, when augmented with a linear \emph{supervised} fine-tuned layer, achieve remarkable accuracy on image classification.  However, the
lightweight alignment step in these methods still requires \emph{labels} mapping raw inputs to class labels.  The core difference of our work is
that we assume there is no external teacher (and therefore no labels) at all once the model is ``deployed''.

Most work on cross-modal alignment assumes parallel data, or simultaneous presence of both modalities.  For example, the work by
\cite{kim2022crossmodal} demonstrates a system that learns aligned embeddings of vision and text.  Like our work, their method exploits the
relational structure (co-occurrence statistics) among both entities in images and among words to inform an ``alignment loss''.  However, this
method assumes simultaneous observation of both modalities to compute the cross-model co-occurrence ($e_{o_i,w_j}$ for visual object $o_i$ and
word $w_j$) required to compute their alignment loss (whereas ours only assumes vision).

Work on alignment of \emph{unparallel} data is also relevant.  For example, \cite{conneau2018word} demonstrates a system that aligns embeddings
of words from two language (e.g., English and Chinese) using a linear model that learns a mapping $W$ between \emph{fixed, pretrained}
embeddings from the respective languages ($X$ and $Y$) to minimize the Frobenius norm between the two embedding spaces $\lVert WX - Y \rVert_F$.
This method avoids mode-collapse by assuming pretrained embeddings.  However, this method isn't directly applicable to our case because our
mapping is from pixels (or vector inputs) to embeddings, making our input ``vocabulary size'' potentially unbounded.  (In our experiments, none
of the ``test'' images were seen during training.)  The even more impressive work of \cite{lample2018unsupervised} extends the previous approach
to translate between sentences, also without parallel corpora, but also relies on initialization using frozen word embeddings.

The authors of \cite{sucholutsky2023gettingalignedrepresentationalalignment} give an in-depth survey and discussion of representation alignment,
and also propose a general framework for aligning representations (Figure 2 in their paper).  This framework suggests using an alignment
function to increase the alignment between two systems.  Our system is atypical for this framework because our two systems (the bigram
distribution and the encoder) take in different inputs (text and images, respectively), our bigram distribution takes no input during training
time, and instead, our alignment function (Equation \ref{eqn:theloss}) uses both the bigram distribution \emph{and} the encoder to make two
predictions for the next character (which is then aligned in a more typical manner).

\section{Discussion} 
We view the core contributions of this paper to be a proof of concept that it's possible for a system to be both modality agnostic and still
have prespecified innate (high level) concepts that are detectable from a freshly trained network.
As compute and data increase exponentially, systems with more plasticity become more expedient than hard-coded systems \cite{sutton2019bitter}.
We hope our contributions in this paper---the formal problem formulation and its solution---will provide the beginnings of a tool that will
allow learning systems to have a great deal of plasticity while still being guided by ``innate'' high-level concepts.

We hope that this paper might lend some insight into broader questions about innateness in intelligent systems.  For example, a drive for
``social acceptance'' seems to be nearly universal in humans \cite{LEARY2022135}.  Where do drives like this come from?  One possibility is that
these drives are \emph{derived} from more basic drives like hunger.  For example, at an early age, a person might observe connections between
social cues and being fed.  Another possibility is that these drives are as innate as imprinting is in birds \cite{mccabe2019visual}.  We hope
that our proof-of-concept helps elucidate how, in principle, the latter may be possible.

\subsection{Future work} 

Since these are early steps, there are several unanswered questions and directions for future work.

The robustness of the approach should be tested with even broader modalities.  The examples in this paper are from permuted CIFAR and permuted
EMNIST, which is hand-written printed characters from the Latin alphabet.  Other related modalities would include other fonts (like cursive) or
a phoneme-based representation (e.g., images of spectrograms of spoken phonemes).

It would also be interesting to see how robust this approach is to domain shift.  The innate component of our system, the bigram table,
is from English Wikipedia, which is also the same dataset used to train the model.  It would be interesting to see how robust our results are
for English from sources other than Wikipedia, or even languages other than English.

In a sense, bigram frequencies encode a basic \emph{relation} among characters that are agnostic to the specific representations of the
characters themselves.  Future work would need to answer how this sort of relational representation could be developed for larger domains, since
the scalability of using an ``innate'' n-gram prior seems limited for such domains.  For example, extending our approach to sentence-level
concepts would require some extensions.  A naive bigram table for a vocabulary of 30,000 \emph{words} is nearly a billion entries.  Can we
replace our n-gram model with a predictive model, for example a pretrained LLM?  One challenge is that our loss assumes our predictive model's
inputs being a (soft) probability distribution instead of discrete tokens.  Sampling or training a model to input probabilities would be
required.

Our loss surface is riddled with local optima, which is unsatisfying and expensive to optimize.  Solving a laughably simple dataset like MNIST
shouldn't require 3 GPU-hours.  It would be more satisfying to either find a related but convex loss space, or use more efficient combinatoric
search methods than the naive random restarts we use.

Inspired by the ``discovery order'', in which successful models tended to correctly classify more-frequent characters first (see Appendix
\ref{appendix:order}), for single-model optimization, we tried a ``curriculum'' learning where the model's first task was to distinguish {\tt
  e}'s from other characters, then {\tt e}'s from {\tt t}'s, etc., but we didn't in succeed reliably finding the global optimum with a single
model.  Further, we tried augmenting the loss function with various methods like a variational loss \cite{kingma2022autoencoding},
pre-processing using principle component analysis, and per-character batch-contrastive loss, etc.\ all without significant improvement.

Finally, our motivating example --a beaver's innate drive to build dams-- involves \emph{actions}, not just detection.  An interesting
longer-term direction is to create \emph{control} systems with ``innate'' drives.  For example, a robot that innately wants to pick up trash.  A
direct extension could simply attach a reward to our {\tt fnord} detector.

\section*{Acknowledgements}
We'd like to thank Aditya Vempathy, Mike Mozer, Ravi Kokku, Ashish Jagmohan, Paul Haley, Tim Oates, Amol Nayate, and Rob
Goldstone for discussions and especially Jeremy Hartman for reviewing several drafts of this paper.

\bibliography{babybeaver}
\appendix
\section{Appendix} 
\label{sec:appendix}

\subsection{Local Optima} 
\label{appendix:optimal}

Single training runs of our model usually got stuck in local optima.  The discussion below is for EMNIST, but CIFAR also had similar issues.

\begin{figure}[ht]
  \centering
  \includegraphics[width=0.5\textwidth]{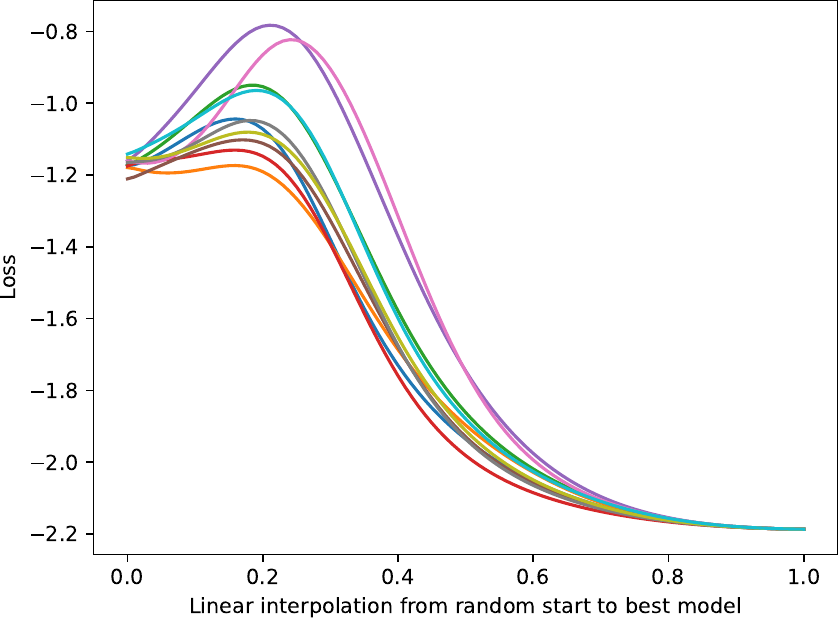}
  \caption{The training loss for linear interpolations between random initializations and the best model found.  Note that the loss is not
    monotonic for most of the seeds.}
  \label{figure:surface}
\end{figure}

We first observed that, when we ``cheat'' and train our encoder with labels using cross-entropy, our loss $\mathcal{L}_f$ is also minimized.
However, gradient descent on our loss often resulted in local optima (far higher than the loss found by ``cheating'', -.7 nats for cheating
vs.\ local optima around -.2 nats vs.\ initial loss of .4 nats), in which the model failed to find a proper correspondence between images and
characters.  This happened with both Xavier \cite{pmlr-v9-glorot10a} and Kaiming \cite{he2015delving} initialization methods, and with different
optimizers (RMS, Adam, SGD).
Following \cite{goodfellow2015qualitatively}, in Figure \ref{figure:surface} we show the training loss as we linearly interpolate the parameters
between several randomly initialized models and the best model we found.  Unlike the loss curves in \cite{goodfellow2015qualitatively}, most of
our curves are not monotonically decreasing, suggesting local optima.  We conjecture that these local optima are due to the combinatoric nature
of the problem: The model is essentially searching for a specific permutation for its mapping from images to indices (classes), without explicit
knowledge (labels) of which ``cluster'' should map to which index.

\subsection{Character Discovery Order} 
\label{appendix:order}

It's worth noting that, during training, the modeled ``discovered'' characters roughly in proportion to their frequency.  That is, during
training, the models tended to first correctly classify the most frequent character {\tt e}, then {\tt a} and {\tt t}.  Below, we show an
example of this on the best (i.e., with lowest final train loss) bigram contrastive model for CIFAR. The ``Characters'' column shows which
characters the model correctly classifies: That is, if the model correctly classifies (choosing the max logit) over half of 100 examples of the
character on an eval set, then we show that character.  Otherwise, we show a ``{\tt .}''.  We also show the step at which point each new
character was ``discovered'''' and the loss at that training step.  Interestingly, {\tt o} is discovered 2000 steps after {\tt h}, despite being
over twice as frequent.  We suspect this is due, in part, to the high frequency of both {\tt t} and the bigram {\tt th}.

\begin{tabular}{lll} 
     {\bf Characters}      &  {\bf Loss} & {\bf Step}
\\ {\tt ....E.....................} & 0.717 & 100
\\ {\tt A...E.....................} & 0.670 &1200
\\ {\tt A...E..............T......} & 0.617 &2300
\\ {\tt A...E.......M......T......} & 0.612 &2400
\\ {\tt A...E.......MN.....T......} & 0.601 &2600
\\ {\tt A...E.......MN...R.T......} & 0.590 &2800
\\ {\tt A...E..H....MN...R.T......} & 0.559 &3300
\\ {\tt A..DE..H....MN...R.T......} & 0.498 &4300
\\ {\tt A..DE..H....MNO..R.T......} & 0.464 &5300
\\ {\tt A..DE..H....MNO..R.TU.....} & 0.451 &5800
\\ {\tt A..DE..H....MNO..RSTU.....} & 0.444 &6100
\\ {\tt A..DE..H...LMNO..RSTU.....} & 0.421 &7400
\end{tabular} 

\subsection{Clustering results} 
\label{appendix:clustering}

\begin{figure}[ht]
  \centering
  \includegraphics[width=0.5\textwidth]{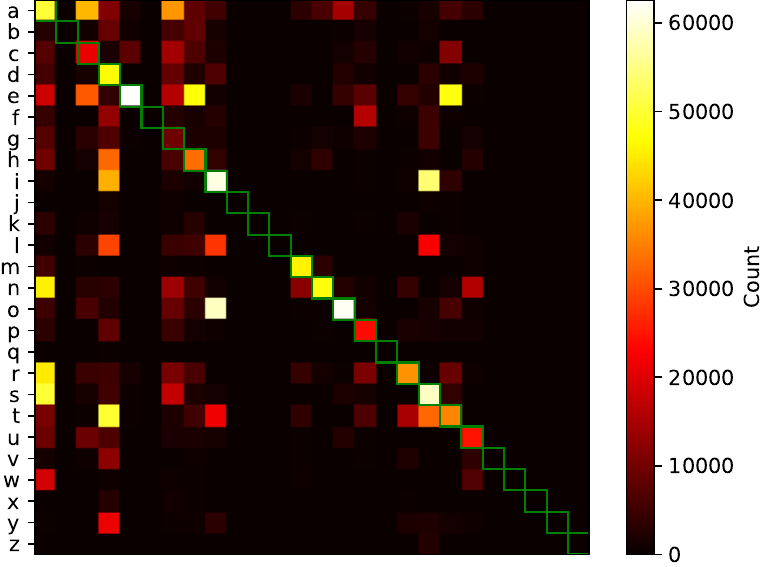}
  \caption{The confusion matrix for clustering with optimized assignments.}
  \label{figure:confusion}
\end{figure}

Here we show the results of K-means clustering showing an absence of correspondence between cluster and the image's true label.  The cluster
assignments are shown in Figure \ref{figure:confusion}.  We assigned the clusters to maximize the trace using the Hungarian algorithm
\cite{kuhn1955hungarian}.  This optimal assignment achieves only 29.44\% total accuracy.  Note that the training data for this method is
weighted by character frequency (i.e., {\tt e} appears more often than {\tt q}), so isn't directly comparable with our method's accuracy on
balanced EMNIST, but max-class accuracy (guessing everything is an {\tt e}) is 11.76\%.

\begin{figure}[ht]
  \centering
  \includegraphics[width=0.5\textwidth]{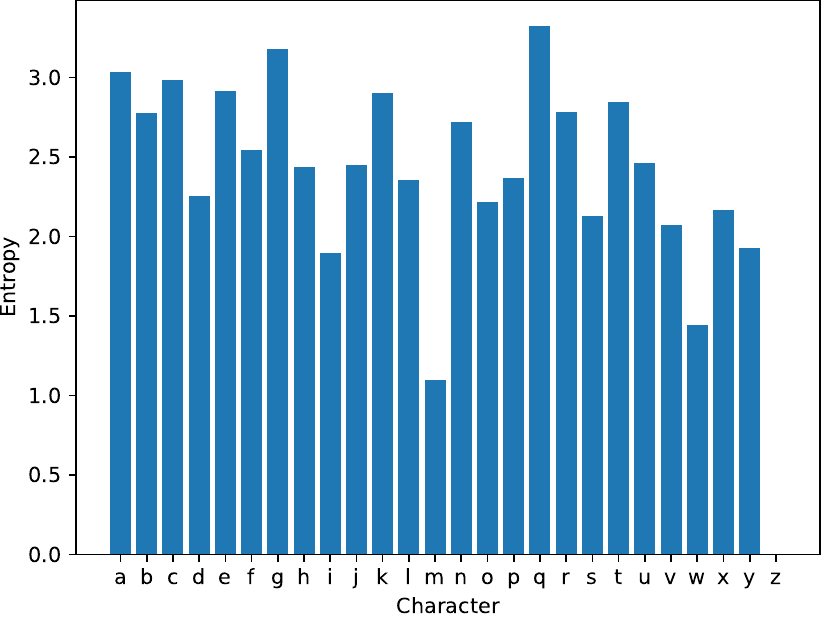}
  \caption{The entropies for the clusters, in nats.}
  \label{figure:entropies}
\end{figure}

The entropies for these clusters are shown in Figure \ref{figure:entropies}.

\subsection{Batch contrastive loss} 
\label{appendix:contrastloss}

Given batch encodings $e$ and $d$ (each with $\l|B\r|$ indices), where $e_{i, j}$ and $d_{i, j}$ are the predicted probability that item $i$ is
class $j$ for $e$ and $d$, respectively, we want to compute the probability that a particular index $i$ gets matched to the same index for $d$.

That is, $e$ defines a probability from indices to classes, $P(c_j|i,e) = e_{i,j}$, from which we derive a probability from classes to indices:
\begin{eqnarray*} 
  P(i|c_j,e) &=& \f{P(c_j|i,e)P(i|e)}{P(c_j|e)}
\\  &=& \f{P(c_j|i,e)}{\l|B\r| P(c_j|e)}
\\  &=& \f{e_{i,j}}{\l|B\r| \f{1}{\l|B\r|}\sum_{k=1}^{\l|B\r|} e_{k,j}}
\\  &=& \f{e_{i,j}}{\sum_{k=1}^{\l|B\r|} e_{k,j}}
\end{eqnarray*} 

Now, we want to know the probability that we can recover the correct index from $d$ given a sample from $e$.
\begin{eqnarray*} 
  P(i_e = i_d) &=& \sum_{j=1}^{\l|\Sigma\r|} P(c_j|i,d) P(i|c_j,e)
\\  &=& \sum_{j=1}^{\l|\Sigma\r|} d_{i,j} P(i|c_j,e)
\\  &=& \sum_{j=1}^{\l|\Sigma\r|} \f{d_{i,j} e_{i,j}}{\sum_{k=1}^{\l|B\r|} e_{k,j}}
\end{eqnarray*} 

Averaging over the negative-log of $P(i_e = i_d)$ for all $i$, we get:
\begin{eqnarray*} 
  \mathcal{L}\l(e, d\r) &=& \f{1}{\l|B\r|} \sum_{i=1}^{\l|B\r|} -\log\l(P(i_e = i_d)\r)
\\ &=& -\f{1}{\l|B\r|} \sum_{i=1}^{\l|B\r|} \log\l(\sum_{j=1}^{\l|\Sigma\r|} \f{d_{i,j} e_{i,j}}{\sum_{k=1}^{\l|B\r|} e_{k,j}}\r)
\end{eqnarray*} 

(In our paper, $d_{i+1}$ is computed from $e_i$ and a distribution over bigram.  Thus, the above loss can be represented as a function of the
encodings $e$ and the ``true'' bigram distribution.  We empirically verified that our loss differs from the KL divergence between the true
bigram distribution, but their exact mathematical relationship remains for future work.)

\subsection{Detectability of Trigger Lengths} 
\label{appendix:triggerlen}

\begin{figure}[ht]
  \centering
  \includegraphics[width=0.5\textwidth]{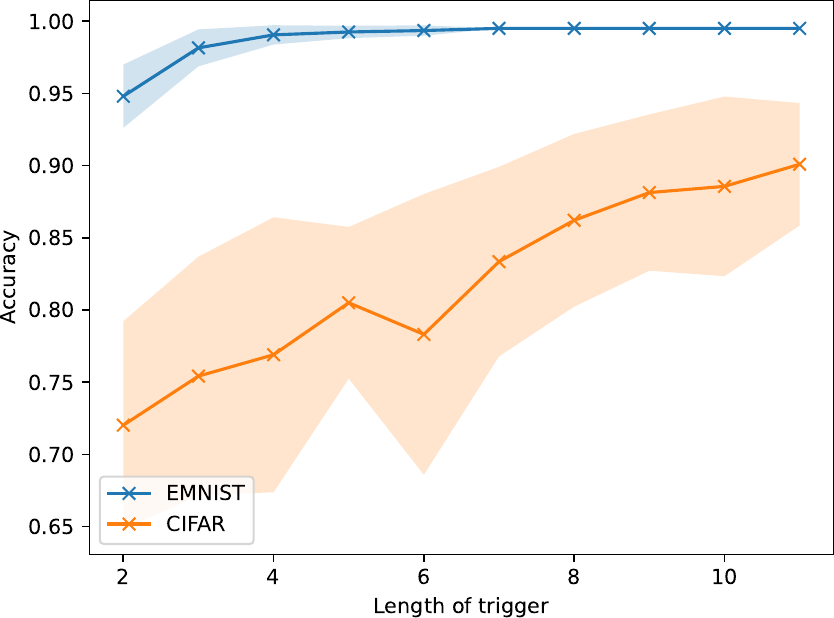}
  \caption{The effect of the detectability of a trigger as a function of trigger length (in characters).  The shaded area represents the
    standard error.}
  \label{figure:triggerlen}
\end{figure}

Figure \ref{figure:triggerlen} shows the effect of trigger detection accuracy (for models trained using the bigram batch contrastive loss) as a
function of trigger length.  Triggers were sampled from the Wikipedia text and also generated as random (uniform) strings, 100 each.  The
Pearson correlation of accuracy to trigger length is .6854 for EMNIST and .9791 for CIFAR.  Generally, longer triggers are easier to detect,
probably due, in part, to the lower likelihood of these triggers occurring by chance.

\subsection{CIFAR 26} 
\label{appendix:cifar}

To test the robustness of our approach to domains outside EMNIST, we reran our entire experimental suite on a ``CIFAR 26'' font.  Instead of
using EMNIST characters for images of letters, the system uses CIFAR100 images for letters, with somewhat arbitrary class-to-letter assignments.
In our initial runs, we simply used the first 26 classes alphabetically.  That is, {\tt a} is represented by images of apples, {\tt b} by
``aquarium fish'', {\tt n} by buses, etc..  The experimental setup is identical to EMNIST with the exception that we changed the encoder's
input dimension to 3,072.  A visual representation of this ``font'' is shown at the bottom of Figure \ref{figure:fnordscore}.

The character to class map is:
\begin{tabular}{rl|rl} 
   {\tt a} & apples         & {\tt n} & bus
\\ {\tt b} & aquarium fish  & {\tt o} & butterfly
\\ {\tt c} & baby           & {\tt p} & camel
\\ {\tt d} & bear           & {\tt q} & cans
\\ {\tt e} & beaver         & {\tt r} & castle
\\ {\tt f} & bed            & {\tt s} & caterpillar
\\ {\tt g} & bee            & {\tt t} & cattle
\\ {\tt h} & beetle         & {\tt u} & chair
\\ {\tt i} & bicycle        & {\tt v} & chimpanzee
\\ {\tt j} & bottles        & {\tt w} & clock
\\ {\tt k} & bowls          & {\tt x} & cloud
\\ {\tt l} & boy            & {\tt y} & cockroach
\\ {\tt m} & bridge         & {\tt z} & keyboard
\end{tabular} 

\end{document}